\documentclass[manuscript]{acmart}

\AtBeginDocument{%
  \providecommand\BibTeX{{%
    \normalfont B\kern-0.5em{\scshape i\kern-0.25em b}\kern-0.8em\TeX}}}

\setcopyright{acmcopyright}
\copyrightyear{2018}
\acmYear{2018}
\acmDOI{10.1145/1122445.1122456}

\acmConference[Woodstock '18]{Woodstock '18: ACM Symposium on Neural
  Gaze Detection}{June 03--05, 2018}{Woodstock, NY}
\acmBooktitle{Woodstock '18: ACM Symposium on Neural Gaze Detection,
  June 03--05, 2018, Woodstock, NY}
\acmPrice{15.00}
\acmISBN{978-1-4503-XXXX-X/18/06}

\usepackage{amsmath}
\usepackage{multirow}
\usepackage{hyperref}
\usepackage{booktabs}
\usepackage{xcolor}

\definecolor{skyblue}{rgb}{0.53, 0.81, 0.92}
\definecolor{BurntOrange}{RGB}{255, 137, 76}
\acmSubmissionID{9674}

\begin{document}

\title{Quda: Natural Language \underline{Q}ueries for Vis\underline{u}al \underline{D}ata \underline{A}nalytics}

\author{Siwei Fu}
\email{fusiwei339@gmail.com}
\affiliation{%
  \institution{Zhejiang Lab}
  \streetaddress{No.1818 Wenyi West Road}
  \city{Hangzhou}
  \state{Zhejiang}
  \country{China}
  \postcode{310000}
}

\author{Kai Xiong}
\email{kxiong@zju.edu.cn}
\affiliation{%
  \institution{Zhejiang University}
  \streetaddress{No.866 Yuhangtang Road}
  \city{Hangzhou}
  \state{Zhejiang}
  \country{China}
  \postcode{310000}
}

\author{Xiaodong Ge}
\email{xdge@zju.edu.cn}
\affiliation{%
  \institution{Zhejiang University}
  \streetaddress{No.866 Yuhangtang Road}
  \city{Hangzhou}
  \state{Zhejiang}
  \country{China}
  \postcode{310000}
}

\author{Siliang Tang}
\email{siliang@zju.edu.cn}
\affiliation{%
  \institution{Zhejiang University}
  \streetaddress{No.38 Zheda Road}
  \city{Hangzhou}
  \state{Zhejiang}
  \country{China}
  \postcode{310000}
}

\author{Wei Chen}
\email{chenvis@zju.edu.cn}
\affiliation{%
  \institution{Zhejiang University}
  \streetaddress{No.866 Yuhangtang Road}
  \city{Hangzhou}
  \state{Zhejiang}
  \country{China}
  \postcode{310000}
}

\author{Yingcai Wu}
\email{ycwu@zju.edu.cn}
\affiliation{%
  \institution{Zhejiang University}
  \streetaddress{No.866 Yuhangtang Road}
  \city{Hangzhou}
  \state{Zhejiang}
  \country{China}
  \postcode{310000}
}


\begin{abstract}

  The identification of analytic tasks from free text is critical for visualization-oriented natural language interfaces (V-NLIs) to suggest effective visualizations. However, it is challenging due to the ambiguity and complexity nature of human language. To address this challenge, we present a new dataset, called Quda, that aims to help V-NLIs recognize analytic tasks from free-form natural language by training and evaluating cutting-edge multi-label classification models. Our dataset contains $14,035$ diverse user queries, and each is annotated with one or multiple analytic tasks. We achieve this goal by first gathering seed queries with data analysts and then employing extensive crowd force for paraphrase generation and validation. We demonstrate the usefulness of Quda through three applications. This work is the first attempt to construct a large-scale corpus for recognizing analytic tasks. With the release of Quda, we hope it will boost the research and development of V-NLIs in data analysis and visualization.
\end{abstract}



\newcommand{\name}{Quda} 
\newcommand{\etal}{et~al.\ }
\newcommand{\eg}{e.g.}
\newcommand{\ie}{i.e.}
\newcommand{\siwei}[1]{\textcolor{blue}{#1}}
\newcommand{\xd}[1]{\textcolor{red}{#1}}
\newcommand{\dattribute}[1]{\colorbox{skyblue}{#1}}
\newcommand{\dvalue}[1]{\colorbox{BurntOrange}{#1}}

\newcommand\Tstrut{\rule{0pt}{2.6ex}}         
\newcommand\Bstrut{\rule[-0.9ex]{0pt}{0pt}}   

\brokenpenalty=100
\tolerance=500
\clubpenalty=10000
\widowpenalty=10000
\exhyphenpenalty=2000
\hyphenpenalty=2000

\setlength{\floatsep}{1mm}
\setlength{\textfloatsep}{2mm}
\setlength{\intextsep}{1mm}
\setlength{\dbltextfloatsep}{2mm}
\setlength{\dblfloatsep}{1mm}

\begin{teaserfigure}
  \includegraphics[width=\linewidth]{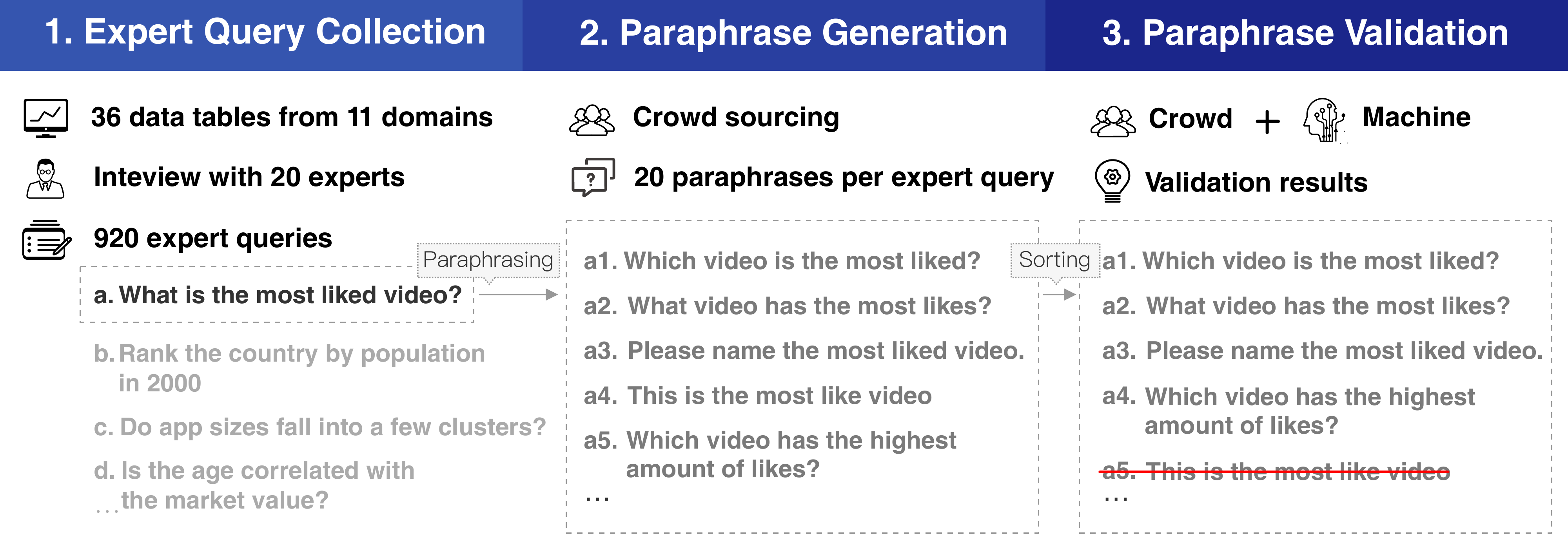}
  \caption{Overview of the entire data acquisition procedure consisting of three stages. In the first stage, we collect $920$ queries by interviewing $20$ data analysts. Second, we expand the corpus by collecting paraphrases using crowd intelligence. Third, we borrow both the crowd force and a machine learning algorithm to score and reject paraphrases with low quality.}
  \label{fig:teaser}
\end{teaserfigure}
\maketitle

\section{Introduction}
Research in visualization-oriented natural language interface (V-NLI) has attracted increasing attention in the visualization community.
Data analytics or dataflow systems such as Articulate~\cite{Sun2010} and FlowSense~\cite{Yu2020} leverage V-NLIs to provide easy-to-use, convenient, and engaging user experience, and facilitate the flow of data analysis.
Existing V-NLIs, however, face two key challenges~\cite{Yu2020}.
On the one hand, making effective design decisions, \eg, choosing a proper visualization, is challenging because of the large design space.
V-NLIs can address this issue by employing findings of empirical studies that provide design recommendations given analytic tasks and datasets~\cite{Saket2019}.
However, addressing this issue requires V-NLIs to distill analytic tasks from queries before making a design choice.

On the other hand, due to the complex nature of human language, understanding free-form natural language input (or ``query'' in the following article) is non-trivial.
Recent research in V-NLI has focused mainly on rule-based language parsers to understand users’ intent~\cite{Yu2020,Srinivasan2018,Setlur2016}.
Although effective, these approaches are narrow in usage scenarios because of limited grammar rules.
Moreover, establishing rules is cumbersome and error-prone, and may require extensive knowledge in both visualization and natural language processing (NLP).

Learning-based approaches have reached state-of-the-art performance in various NLP tasks and have great potential in understanding free-form queries in V-NLIs. 
Large-scale training examples catered to visual data analytics are critical to developing and benchmarking learning-based models to be deployed for V-NLIs.
For example, multi-label text classification techniques trained with large amount of query-task pairs are able to categorize user queries into analytic tasks, which contributes to the choice of proper visualization. 
Admittedly, the acquisition of such a corpus is difficult for two reasons.
First, the target users of V-NLIs are often identified as data analysts who are knowledgeable in analytic tasks~\cite{Srinivasan2018,Yu2020}. 
Therefore, a high-quality corpus should reflect how analysts interact with data under various tasks.
Second, compared to general-purpose datasets, such as existing corpora collected via social media~\cite{Lan2017} or online news corpora~\cite{Zhang2015}, expert queries for V-NLIs are not naturally occurring. 



In this paper, we present a textual corpus, \name{}\footnote{We make our corpus and related materials publicly available on \url{https://freenli.github.io/quda/}}, that is the first trail to bridge learning-based NLP techniques with V-NLIs from the perspective of dataset.
\name{} consists of a large number of query-task pairs, aiming to train and evaluate deep neural networks to categorize queries into analytic tasks.
Our corpus features high-quality and large-volume queries designed in the context of V-NLIs.
To accomplish this goal, we first design a lab study to collect seed queries by interviewing data analysts.
Next, to augment data volume and diversify the ways of saying something, we employ the crowd to collect a number of sentential paraphrases for each query. 
These queries are restatements with approximately the same meaning.
Finally, we design a validation procedure for quality control.
The \name{} dataset contains $920$ utterances created by data analysts, and each are accompanied by $14$ paraphrases on average.
All queries are annotated with at least one of ten analytics tasks.

To present the usefulness of \name{}, we run three experiments. 
In the first experiment, we train and evaluate a multi-label classification model on \name{} with different data splits. 
We reveal that \name{} is a challenging dataset for multi-label classification, and leaves room for developing new classification models for V-NLIs.
Second, we use the model trained on \name{} to augment an existing V-NLI toolkit, showing that the model is more robust in task inference.
Third, with the help of \name{}, we build a query recommendation technique that enables V-NLIs to suggest valuable queries given partial or vague user input.
To summarize, the primary contributions of this paper are as follows:

\begin{itemize}
    \item By employing both the expert and crowd intelligence, we construct a large-scale and high-quality corpus, named \name{}, that contains $14,035$ queries in the domain of visual data analytics.
    \item Building upon existing research on V-NLIs and task taxonomies, we define the space of queries for V-NLIs and highlight the scope of our corpus from six characteristics.
    \item We conduct three experiments to show that \name{} can benefit a wide range of applications, including 1) training and benchmarking deep neural networks, 2) improving the robustness of task inference for V-NLIs, and 3) providing query recommendation for V-NLIs.
    
\end{itemize}

With the release of \name{}, we hope it will stimulate the publication of similar large-scale datasets, and boost research and development of V-NLIs in data analysis and visualization.

\section{Related Work}

\subsection{NLI for Data Visualization}
Our research is motivated by recent progress in V-NLIs, which provides engaging and effective user experience by combining direct manipulation and natural language as input.

Some research emphasizes on the multi-modal interface that interacts with users using natural language.
Back in 2010, Articulate~\cite{Sun2010} identifies nine task features and leverages keyword-based classification methods to understand how user's imprecise query weighted in each feature.
Sisl~\cite{Cox2001} is a multi-modal interface that accepts various input modes, including point-and-click interface, NL input in textual form, and NL using speech.
Analyza~\cite{Dhamdhere2017} is a system that combines V-NLI with a structured interface to enable effective data exploration.
To address the ambiguity and complexity of natural language, Datatone~\cite{Gao2015} presents a mixed-initiative approach to manage ambiguity in the user query.
Flowsense~\cite{Yu2020} is a V-NLI designed for the dataflow visualization system.
It applies semantic parsing to support natural language queries and allows users to manipulate multiple views in a dataflow visualization system.
Several commercial tools, such as Microsoft Power BI~\cite{microsoft}, IBM Watson Analytics~\cite{IBM}, Wolfram Alpha~\cite{wolfram}, Tableau~\cite{tableau}, and ThoughtSpot~\cite{thoughtspot}, integrate V-NLIs to provide better analytic experience for novice users. 

Another line of research targets the conversational nature of V-NLI.   
Fast \etal proposed Iris~\cite{Fast2018}, a conversational user interface that helps users with data science tasks.
Kumar \etal~\cite{Kumar2016} aimed to develop a data analytics system to automatically generate visualizations using a full-fledged conversational interface.
Eviza~\cite{Setlur2016} and Evizeon~\cite{Hoque2018} are visualization systems that enable natural language interactions to create and manipulate visualizations in a cycle of visual analytics. 
Similarly, Orko~\cite{Srinivasan2018} is a prototype visualization system that combines both natural language interface and direct manipulation to assist visual exploration and analysis of graph data.

The aforementioned approaches are mainly implemented using rule-based language parsers, which provide limited support for free-form NL input.
Instead of generating visualizations, Text-to-Viz~\cite{Cui2020} aims to generate infographics from natural language statements with proportion-related statistics. 
The authors sampled $800$ valid proportion-related statements and built a machine learning model to parse utterances.
Although promising, Text-to-Viz does not support queries for a broad range of analytic activities.

The performance and usage scenario of V-NLIs highly depend on language parsers.
The cutting-edge NLP models have reached close-to-human accuracies in various tasks, such as semantic parsing, text classification, paraphrase generation, etc. 
However, few have been applied to visualization-oriented V-NLIs.
We argue that the release of the high-quality textual dataset would assist the training and evaluation of NLP models in the domain of data analytics.

\subsection{Datasets for visualization research}

An increasing number of studies in the visualization community have employed supervised learning approaches in various scenarios such as mark type classification~\cite{Savva2011}, reverse engineering~\cite{Poco2017}, color extraction, etc. 
The capability of these approaches, however, highly relies on the availability of massive datasets that are used for training and evaluation~\cite{Lee2019}. 

Some datasets consisting of visual diagrams are constructed and open-sourced for visualization research. 
For example, Savva \etal~\cite{Savva2011} compiled a dataset of over 2,500 chart images labeled by chart type. 
Similarly, the Massvis dataset~\cite{Borkin2013} contains over 5,000 static chart images, and over 2,000 are labeled with chart type information.
Beagle~\cite{Battle2018} embeds a Web Crawler extracting more than 41,000 SVG-based visualizations from the web, and all are labeled by visualization type.
Viziometrics~\cite{Lee2018} collects about 4.8 million images from scientific literature and classifies them into five categories, including equation, diagram, photo, plot, and table.
Poco and Heer~\cite{Poco2017} compiled a chart corpus where each chart image is coupled with bounding boxes and transcribed text from the image.
Instead of compiling a corpus containing chart images, \name{} includes queries accompanied by analytic tasks for visualization-oriented V-NLIs.

Viznet, on the other hand, collects over 31 million real-world datasets that provides a common baseline for comparing visual designs.
Though containing rich textual information, Viznet can hardly be applied to provide training samples for learning-based NLP models.
On the contrary, \name{} is designed for helping V-NLIs understand utterances describing analytic tasks in information visualization.

\subsection{Corpora for Multi-label Text Classification}
\begin{table}[ht]
    \centering
    \begin{tabular}{l|ccccc}
    \toprule
    & \textbf{\# Instances} & \textbf{\# Categories} & \textbf{Cardinality} & \textbf{Density} & \textbf{Diversity} \\\midrule
    20 NG   & 19,300       & 20            & 1.029       & 0.051   & 0.003     \\
    Ohsumed & 13,930       & 23            & 1.663       & 0.072   & 0.082     \\
    IMDB    & 120,900      & 28            & 2.000       & 0.071   & 0.037     \\
    TMC2007 & 28,600       & 22            & 2.158       & 0.098   & 0.047     \\
    Quda    & 14,035       & 10            & 1.331       & 0.133   & 0.054     \\
    \bottomrule
    \end{tabular}
    \caption{Compare \name{} with four well-known multi-label text classification corpora on five criteria, including the number of instances, the number of categories, cardinality, density, and diversity.}
    \label{tab:multilabel}
\end{table}

Our corpus is primarily designed for helping V-NLIs classify queries into multiple analytic tasks. 
Therefore we survey related corpora for multi-label text classification, which is the task of categorizing a piece of textual content into $n$ possible categories, where $n$ can be $0$ to $n$ inclusive.
The website~\cite{multilabel} lists a large number of datasets for multi-label text classification. 
We selectively introduce four datasets that are comparable to \name{}.
For example, the 20 NG or 20 Newsgroups dataset~\cite{Lang1995} includes about $20,000$ documents collected and categorized from $20$ different newsgroups.
Ohusumed~\cite{Joachims1998} contains $13,930$ medical abstracts belonging to $23$ cardiovascular disease categories.
The IMDB dataset contains $120,919$ text summaries of the movie plot, and each is labeled with one or more genres.
The TMC2007 dataset contains $28,569$ aviation safety reports that are associated with $22$ problem types.
Different from the aforementioned datasets, \name{} is the first corpus targeting queries in V-NLIs.

Table~\ref{tab:multilabel} shows the statistical comparison between \name{} and the aforementioned datasets.
Inspired by the prior work~\cite{Moyano2017}, we use five metrics to characterize each dataset. 
Besides the number of instances and categories, we also compute cardinality (\ie, the average number of categories for each instance), density (\ie, the cardinality divided by the number of categories), and diversity (\ie, the percentage of category sets present in the dataset divided by the number of possible category sets.)
Detailed definitions and calculations are described in~\cite{Moyano2017}.
We notice that \name{} is comparable with other corpora in most metrics.
Especially, \name{} has the highest density among all corpora, which means instances in \name{} have higher probability of being assigned with multiple categories.

\section{Problem Definition}
\label{sec:def}
Analytic tasks play a pivotal role in visualization design and evaluation.
The understanding of tasks is beneficial for a wide range of applications in the visualization community such as visualization recommendation, computational linguistics for data analytics, etc.
The goal of this paper is to propose a corpus in the domain of visual data analytics that facilitates the deployment of learning-based NLP techniques for V-NLIs. 
Specifically, our corpus focuses on queries that reflect how data analysts ask questions in V-NLIs.
Due to the variation and ambiguity of natural language, the space of queries is intangibly large.
To narrow down the scope of our corpus, we borrow the idea from prior work in visualization framework~\cite{Munzner2014,Amar2005,Rind2016,Srinivasan2017} and V-NLI for visualization~\cite{Tory2019,Srinivasan2018} and identify \name{} from six characteristics, \ie, abstraction level, composition level, perspective, type of data, type of task, and context-dependency.

\textbf{Abstraction: Concrete.}
The abstraction level is one characteristic of analytic tasks describing the concreteness of a query~\cite{Rind2016}.
Abstract queries are generic and are useful for generalizing the concept behind a concrete task (an example is, \textit{``Find maximum''}).
These queries can be addressed in multiple ways based on the interpretation.
However, to obtain a reasonable response from V-NLIs, an analyst should author a query that provides sufficient information~\cite{Srinivasan2018}, such as \textit{``Find top movies with most stars''} and \textit{``Retrieve the country with the most population}''.
Therefore, our corpus focuses on the queries at the low abstraction level which expresses both tasks and values explicitly.

\textbf{Composition: Low.}
The composition level describes the extent to which a query encompasses sub-queries~\cite{Rind2016}.
Composition is a continuous scale from low-level to high-level.
A query with a high composition level consists of multiple sub-queries and a V-NLI may answer it using multiple steps.
For example, \textit{``For the movie with the most stars, find the distribution of salaries of the filming team.''}
A V-NLI first needs to identify the movie with the most stars. 
Then display the salary distribution of all staff in the filming team.
As our corpus is the first attempt to collect queries for V-NLI, we focus on queries that are low in the composition level at this stage. 
We plan to include with high composition level in future research.

\textbf{Perspective: Objectives.}
Queries in V-NLIs can be classified into two categories, \ie, objectives and actions~\cite{Rind2016}.
Objectives are queries raised by analysts seeking to satisfy a curiosity or solve a problem.
The actions in V-NLIs are executable steps towards achieving objects, and usually relate to interactive features of visualization artifacts, such as \textit{``Show the distribution using a bar chart,''} or \textit{``Map the popularity of applications to x-axis.''}
However, to assume V-NLI users being knowledgeable in constructing effective visualization is virtually arbitrary.
Instead of enumerating actions, we aim to collect queries that are objectives raised by data analysts.

\textbf{Type of Data: Table.}
The type of dataset affects the syntax and semantics of queries.
For example, analysts may seek to identify the shortest path between two nodes for a network dataset.
However, such query rarely occurs in the tabular dataset because links between items are not supported explicitly.
Data and dataset can be categorized into five types~\cite{Munzner2014}, including table, networks \& trees, fields, geometry, and clusters \& sets \& lists.
At the date of paper submission, our corpus target at queries based on the tabular dataset, we argue that our corpus can support other types of data to some extent.
For example, the task of ``Find Nodes'' in networks is similar to ``Retrieve Value'' in tabular data.
We plan to extend our research to support other data types comprehensively in the future.

\textbf{Type of Tasks: 10 Low-level Tasks.}
Collecting a corpus for all possible tasks is not feasible.
Therefore, we turned to the related studies to identify analytic tasks on which to focus.
In this study, we adopt the taxonomy proposed by Amar \etal~\cite{Amar2005} that categorized ten low-level analytic activities, such as Retrieve Value, Sort, etc. 
These tasks serve a good starting point for our research despite not being comprehensive.

\textbf{Context Dependency: Independent.}
The conversational nature of V-NLI may result in queries that rely on contextual information~\cite{Srinivasan2018,Tory2019}.
For example, a query \textit{``Find the best student in the class''} may be followed by \textit{``Obtain the English score of the student.''}
The second query appears to be incomplete individually and is a follow-up query of the first query. 
Such a query is referred to as contextual queries, which is not the focus of our research.
Our corpus focuses on queries that have complete references to tasks or values associated with a task.

\section{Constructing \name{}}
In this work, we incorporate both expert and crowd intelligence in data acquisition, which can be divided into three stages.
In the first stage, we conduct an interview study involving $20$ data analysts to author queries based on $36$ data tables and $10$ low-level tasks.
We derive $920$ expert queries from the study results. 
Some example queries are shown in Figure~\ref{fig:teaser} (Expert Query Collection).
For the second stage, we perform a large-scale crowd-sourced experiment to collect sentential paraphrases for each expert query, which are restatements of the original query with approximately the same meaning~\cite{Wang2018}.
Example crowd-crafted queries are shown in Figure~\ref{fig:teaser} (Paraphrase Generation). 
That is, for an expert sentence \textit{``what is the most liked video?''}, five paraphrases  (a1---a5) are crafted by the crowd.
Finally, we design a validation stage to estimate the semantic equivalence score for crowd-generated queries, and accepted those with high scores.
The third stage results in $13,115$ paraphrases where each expert sentence is accompanied with $14$ paraphrases on average.
An example is depicted in Figure~\ref{fig:teaser} (Paraphrase Validation), the five queries are re-ordered and the last one is filtered out due to low semantic similarity.

\section{Employing Expert Intelligence}
In this section, we describe expert interviews that aim to collect professional queries from data analysts, given analytic tasks and data tables in different domains.

\subsection{Participants and Apparatus}
To understand how data analysts raise questions from data tables, we recruit 20 individuals, 10 females, and 10 males, with ages ranging from 24---31 years ($\mu =25.45, \sigma =2.80$). 
In our study, we identify ``data analysts'' as people who are experienced in data mining or visual analytics and have at least one publication in related fields.
Most participants ($16/20$) are postgraduate students majoring in Computer Science or Statistics, and the rest are working as data specialists in IT companies.
All of them are fluent in English and use it as working language.
The experiments are conducted on a laptop (2.8GHz 4-Core Intel Core i7, 16 GB memory) on which participants read documents and create queries.

\subsection{Tasks and Data Tables}
Tasks play a vital role in authoring queries.
Our study begins with the taxonomy of $10$ low-level visualization tasks~\cite{Amar2005}, including Retrieve Value, Compute Derived Value, Find Anomalies, Correlate, etc.
The pilot study shows that participants may be confused about some tasks.
For example, one commented, ``\textit{Determine Range is to find the maximum and minimum values in one data field, which is similar to Find Extremum.}''
To help participants clarify the scope of each task, we compiled a document presenting each task from three aspects, \ie, general description, definition, and example sentences.
The document is shown in the supplementary material.

Data tables provide a rich context for participants to create queries.
Hence, to diversify the semantics and syntax of queries, we have prepared $36$ real-world data tables covering $11$ different domains, including health care, sports, entertainment, etc. 
Tables with insufficient data fields may not support some types of queries.
For example, assume a table about basketball players has two columns, \ie, players' name and their nationality. 
Participants may find it hard to author queries in the ``Find Extremum'' or ``Correlate'' categories which usually require numeric fields.
Hence, instead of skipping the tasks, we allow participants to revise tables by adding new columns or editing existing ones if necessary.
Moreover, we selectively choose data tables that have rich background information and explanation for each column so that participants can get familiar with them in a short time.
All data were collected from Kaggle~\footnote{\url{https://www.kaggle.com/datasets}}, Data World~\footnote{\url{https://data.world/}}, and Google Datasetsearch~\footnote{\url{https://datasetsearch.research.google.com/}}. 

\subsection{Methodology and Procedure}
The interview began with a brief introduction to the purpose of the study and the rights of each participant.
We then collected the demographic information of each participant, such as age, sex, experience in data mining and visual analytics.
After that, participants were asked to familiarize themselves with $10$ analytic tasks by reading the document. 

In the training stage, participants were asked to author queries for $10$ tasks based on a sample data table, which differs from those used in the main experiment.
We instructed them to think aloud and resolved any difficulties they encountered.
We encourage participants to author queries with diverse syntax structure and semantics. 

The pilot results indicate that the participants were more enthusiastic in generating sentences with diverse syntax if they are interested in the context of the table.
At the beginning of the main experiment, we presented $10$ tables with a detailed explanation of the context and data fields to participants and encouraged them to choose two based on their interest.
We randomized the presentation order of the $10$ tasks.
Participants were guided to author at least two queries given a table and a task, and no time limit was given to complete each query. 
To summarize, the target queries collected at this stage is $20 (participants)\times 2 (tables) \times 10 (tasks) \times 2 (queries)=800 (queries)$
However, because some of the participants enjoyed the authoring experience and generated more sentences for some tasks, the total number of queries exceeded $800$.

After the main experiment, participants were asked to re-assign task labels for each query. 
Because some queries belong to multiple tasks in nature. 
For example, The sentence \textit{``For applications developed in the US, which is the most popular one?''} falls in both \textit{``Filter''} and \textit{``Find Extremum''} categories.
The identification of all task labels is important to reflect the characteristics of each query.

Finally, a post-interview discussion was conducted to collect their feedbacks.
Interviews were audio recorded and we took notes of the participants' comments for further analysis. 
Each interview lasted approximately two hours, and we sent a $\$15$ gift card to interviewees for their participation.

\subsection{Results}
The result of our interview study is a corpus containing $920$ queries generated by data analysts. 
The length of queries ranged from $4$ to $35$, with an average of $11.9$ words ($\sigma =3.9$).

We observed that participants found it enjoyable to write queries for these tasks. 
For example, one participant commented in the post-interview discussion, \textit{``Find Anomalies is an interesting task because it requires an in-depth understanding of the data table.''}
She further added, \textit{``I asked `Based on the relationship among rating, installs, and the number of reviews, which app's rating does not follow the trend?' because this kind of abnormal apps is worth investigating.''}

\section{Borrowing Crowd Intelligence}

Given a corpus with $920$ queries, the language features are limited.
To construct a high-quality corpus that reflect the ambiguity and complexity of human language, we extend our corpus by collecting paraphrases for expert sentences using crowd intelligence.
This stage can be divided into two steps, \ie, paraphrase generation and validation.
All experiments were conducted on MTurk. 

\subsection{Paraphrase Acquisition at Scale}
In this step, we aim to acquire at least $20$ paraphrases for each expert sentence. 
We followed the crowdsourcing pipeline~\cite{Burrows2013} to collect paraphrases using crowd intelligence. 
We developed a web-based system that the crowd can access through MTurk. 
Our system records both crafted queries and user behavior, \eg, keystrokes and duration, for further analysis.

We describe our task as~\cite{Potthast2010}, \textit{``Rewrite the original text found below so that the rewritten version has the same meaning, but a completely different wording and phrasing''} to help crowd workers understand our task. 
The interface also encourages the crowd to craft paraphrases that differ from the original one in terms of sentence structure.
We demonstrate both valid and invalid examples to explain our requirements.
The interface displays an experts' sentence and instructs workers to rewrite it.
After a sentence is submitted, 
workers are allowed to author paraphrases for other expert sentences, skip sentences, or quit the interface by clicking ``Finish''.
Finally, we use an open-ended question to collect workers' comments and suggestions for this job.
We sent 0.1\$ as a bonus for each sentence after validation.

Based on prior work in corpus collection~\cite{Lasecki2013,Burrows2013} and the pilot experiments, we established a set of rules in the main experiment to ensure the quality of the collected sentences.
First, to collect paraphrases that are diverse in structure, we limited the maximum number of sentences one could create ($20$ in our study) to involve more crowd intelligence.
Second, to obtain sentences with high-quality, workers were informed that the results would be reviewed and validated before they receive a bonus.
Third, the interface recorded and analyzed their keystrokes and duration in crafting sentences to reject invalid inputs. 
For example, we rejected sentences crafted within $5$ seconds or with less than $10$ keystrokes.
Fourth, to avoid duplicate sentences with other workers, we compared the input with all sentences in the database.

\subsection{Validation via Pairwise Comparison}

The result of the first step is a corpus containing $18,400$ crowd-generated paraphrases accompanied by $920$ expert-generated sentences.
The goal of this step is to filter out paraphrases that are semantically inequivalent compared to the expert sentence. 

Due to the ambiguity in human language, asking workers directly to identify whether paraphrases are semantically equivalent to the expert sentence would be restrictive and unreliable. 
For example, given an expert's sentence, \textit{``How many faculties are there at Harvard University?''}, and a worker's, \textit{``How many departments are there at Harvard?''}, it is not clear whether the two sentences have the same meaning because ``faculty'' could mean both ``department'' and ``professor''.
Instead of asking whether a paraphrase is semantically equivalent, we asked comparison questions to capture the equivalence strength of one paraphrase with respect to others.
To be specific, given an expert sentence, a worker is shown a pair of paraphrases written by the crowd and asked to choose which one has closer meaning to the expert's sentence. 
Pairwise comparison is effective in capturing semantic relationships~\cite{Parikh2011}. 
However, exhaustive comparisons, \ie, each paraphrase compares to all other paraphrases using a large number of workers, is unnecessary and cost-prohibitive.
Hence, borrowing from the idea of relative attributes~\cite{Parikh2011}, we design a strategy to collect pairwise comparison dataset for all paraphrases, and train scoring functions to estimate the semantic equivalence scores of them.




\subsubsection{Collecting Pairwise Dataset} 
Each expert sentence is associated with $20$ paraphrases, and thus, we collect pairwise data among them.
The total number of comparisons for each expert sentence is $\frac{m\cdot n}{2}$, where $m$ is the number of paraphrases, and $n$ is the number of pairwise comparisons per paraphrase.
In our experiment, $m=20$, and $n=5$. 
To alleviate randomness, each comparison is completed by $5$ unique workers, providing $\frac{20\cdot 5}{2}\cdot 5=250$ comparison tasks for each expert sentence.
Given $920$ expert sentences, this produces a $250\times 920=230,000$ tasks for $6,304$ unique workers.

We set up a web-based system to assist crowd workers in conducting comparison jobs.
The system displays job descriptions, examples and requirements. 
Each job consists of $10$ comparison tasks based on distinct expert sentences, along with two gold standard instances for quality control.
We manually crafted a set of gold standard tasks that are unambiguous and have a clear answer.
We rejected any jobs where one of the gold tasks was failed. 
To measure subtle differences in the semantics, we used a two-alternative forced-choice design for each comparison in which workers cannot answer ``no difference''.
The workers were paid \$0.15 per job.

\subsubsection{Estimating Semantic Equivalence Score}
In this step, we aim to estimate the semantic equivalence scores for each paraphrase and filter out those that are low in the score.
After collecting a set of pairwise comparison data, we employed relative attribute~\cite{Parikh2011} to learn ranking functions which estimate the strength of the semantic equivalence of a paraphrase with respect to other paraphrases.

Given an expert sentence $e$, we had $20$ training paraphrases $\{p_i\}$ represented by feature vectors $\{x_i\}$ in $\mathbb{R}^n$.
We have collected a set of ordered pairs of paraphrases $O=\{(p_i, p_j)\}$ such that $(p_i, p_j)\in O \Longrightarrow p_i\succ p_j$, \ie, paraphrase $i$ is more close to the expert sentence $e$ in semantics than $j$.
Our goal was to learn a scoring function:
\begin{equation}
    \label{eq:score}
s_e(x_i)=w^\intercal x_i
\end{equation}
such that the maximum number of the following constraints is satisfied:
\begin{equation}
\forall (p_i, p_j)\in O:w^\intercal x_i>w^\intercal x_j
\end{equation}
To capture semantic features of each paraphrase, we employed Universal Sentence Encoder~\cite{Cer2018} and obtained a $512$-dimension feature vector $x_i$ for each paraphrase.
We used a SVM-based solver~\cite{Parikh2011} to approximate $w^\intercal$, and obtained the estimated score for $20$ paraphrases using Equation~\ref{eq:score}.

We set a threshold $T$, so that paraphrases with scores lower than $T$ are filtered out.
The scores range from $-0.8$ to $1.05$, and we set $T=0$ to reject all paraphrases with negative scores.
Finally, we obtained $13,115$ high-quality paraphrases for $920$ expert sentences.

\section{\name{} Applications}

We present three cases showing that \name{} can benefit the construction of V-NLIs in various scenarios. 
The first case demonstrates the characteristics of \name{} by training and benchmarking a well-performing multi-label classification model. 
We then present an example to showcase how \name{} can improve an existing NLI toolkit in task inference.
Finally, we explore how \name{} can assist query recommendation in a V-NLI interface.

\subsection{Multi-label Text Classification}
\label{subsec:multilabel}
Assigning all task labels to a query is critical to V-NLIs.
Therefore, we conduct an experiment to evaluate how a well-performed multi-label classification model performs on our corpus, and reveal the characteristics of our corpus with different experiment settings.

\subsubsection{Model}
Recent research in pre-trained models~\cite{Devlin2019,Yang2019} has achieved state-of-the-art performance in many of the NLP tasks.
These models are trained on large corpora, \eg, all Wikipedia articles, and allow us to fine-tune them on downstream NLP tasks, such as multi-label text classification.
Here, we conduct our experiment with a well-known pre-trained model, Bert, which obtained new state-of-the-art results on eleven NLP tasks when it was proposed~\cite{Devlin2019}.
We trained the model for $3$ epochs, and use a batch size of $2$ and sequence length as $41$, \ie, the maximum possible in our corpus. 
The learning rate was set as $3e-5$, the same as in the original paper.

\subsubsection{Dataset and Evaluation Metrics}
Different data splits (train and test) may affect the performance of models. 
We split \name{} in two ways. First, we randomly split it into the train ($11228$ samples) and test ($2807$ samples), and name it as \name{}\_random.
Second, we split the dataset by experts (\name{}\_expert). 
That is, 20 experts are split into the train ($16$ experts) and test sets ($4$ experts), and all queries created by the same experts and their paraphrases are in the same set.
We obtain $11065$ samples for training and $2970$ samples for test.
Following the work~\cite{Wang2016}, we employ precision and recall as evaluation metrics.

\subsubsection{Results}
The precision and recall of the model on \name{}\_random are $97.7\%$ and $96.0\%$, respectively.
On the other hand, the model performance trained on \name{}\_expert only achieves $84.2\%$ (precision) and $74.3\%$ (recall).
Since the train and test samples of \name{}\_expert are crafted based on different data tables and experts, the syntax space may differ. 
As a result, \name{}\_expert is more challenging. 
Though models trained on \name{}\_random have performed better, the performance on \name{}\_expert is close to real-world scenarios since the data tables and users may vary.
This leaves room for developing new methods that are specifically designed for this task. 

To understand how the model on \name{}\_expert performs across different analytic tasks, we investigate the per-class precision and recall.
As shown in Table~\ref{tab:perclass}, the model performs well on \textit{Find Extremum}, \textit{Sort}, and \textit{Cluster}.
However, it performs poorly in \textit{Filter} and \textit{Characterize Distribution}.
In some tasks, such as \textit{Compute Derived Value} (precision: 0.82, recall: 0.51) and \textit{Retrieve Value} (precision: 0.92, recall: 0.64), the model is high in precision while low in recall.
This means the model returns fewer results compared to the ground-truth labels, but most of the predicted labels are correct.
The understanding of why the performance varies across analytic tasks is beyond the scope of our paper. 
We leave the problem as future research which involves a deep understanding of deep learning models as well as syntax space of different tasks.


\begin{table}
    \centering
    \begin{tabular}{l|cc}
    \toprule
    \textbf{Task}             & \textbf{Precision} & \textbf{Recall} \\\midrule
    Retrieve Value            & 0.92      & 0.64   \\
    Filter                    & 0.79      & 0.79   \\
    Compute Derived Value     & 0.82      & 0.51   \\
    Find Extremum             & 0.86      & 0.88   \\
    Sort                      & 0.94      & 0.81   \\
    Determine Range           & 0.9       & 0.7    \\
    Characterize Distribution & 0.73      & 0.69   \\
    Find Anomalies            & 0.82      & 0.75   \\
    Cluster                   & 0.88      & 0.91   \\
    Correlate                 & 0.9       & 0.76   \\
    \bottomrule
    \end{tabular}
    \caption{The per-class precision and recall of Bert on \name{}\_expert.}
    \label{tab:perclass}
\end{table}

\subsection{Improving an Existing V-NLI Toolkit}
The multi-label classification model trained with \name{} can benefit existing NLI toolkits in task inference for free-form natural language.
We choose NL4DV~\cite{Narechania2020} as an example, which is the first python-based library supporting prototyping V-NLIs.
NL4DV takes NL queries and a dataset as input, and returns an ordered list of Vega-Lite specifications to respond to these queries.
Its query interpretation pipeline includes four modules, such as query parsing, attribute inference, task inference, and visualization generation.
The task inference module of NL4DV focuses on the detection of 1) task type, 2) the parameters associated with a task, and 3) ambiguities of operators and values.
The toolkit currently supports five base tasks (\ie, \textit{Correlation}, \textit{Distribution}, \textit{Derived Value}, \textit{Trend}, \textit{Find Extremum}), and \textit{Filter}.
To infer tasks, the authors set rules according to task keywords, POS tags, dependency tree, and data values.
Although effective in some scenarios, the rule-based approach is hard to extend and limited for free-form natural language.

We aim to improve the precision and range of task type inference with the help of \name{}.
We first use Bert~\cite{Devlin2019} trained on \name{}\_expert (details are introduced in Section~\ref{subsec:multilabel}) to detect task type.
Our model supports ten low-level tasks and can assign multiple task labels for each query. 
Though our model cannot explicitly detect the \textit{Trend} task supported by NL4DV, \textit{Trend} is usually identified as \textit{Correlation} and \textit{Distribution}, and appropriate visualizations are recommended.
For example, \textit{``Illustrate the distribution of the population of Ashley across the five years.''} is identified as \textit{Distribution} and \textit{Filter}. 
Second, we setup a set of rules to generate visualization for all tasks.
Similar to NL4DV, the ten tasks are divided into $9$ base tasks and \textit{Filter}.
For base tasks, we take into account the type of data attributes and tasks to identify proper visualizations and visual encodings.
For \textit{Filter} and its associated attributes, we highlight information in the Bar chart rather than filter to few bars according to the design guideline~\cite{Hearst2019}.

\begin{figure}[tb]
    \centering
    \includegraphics[width=\linewidth]{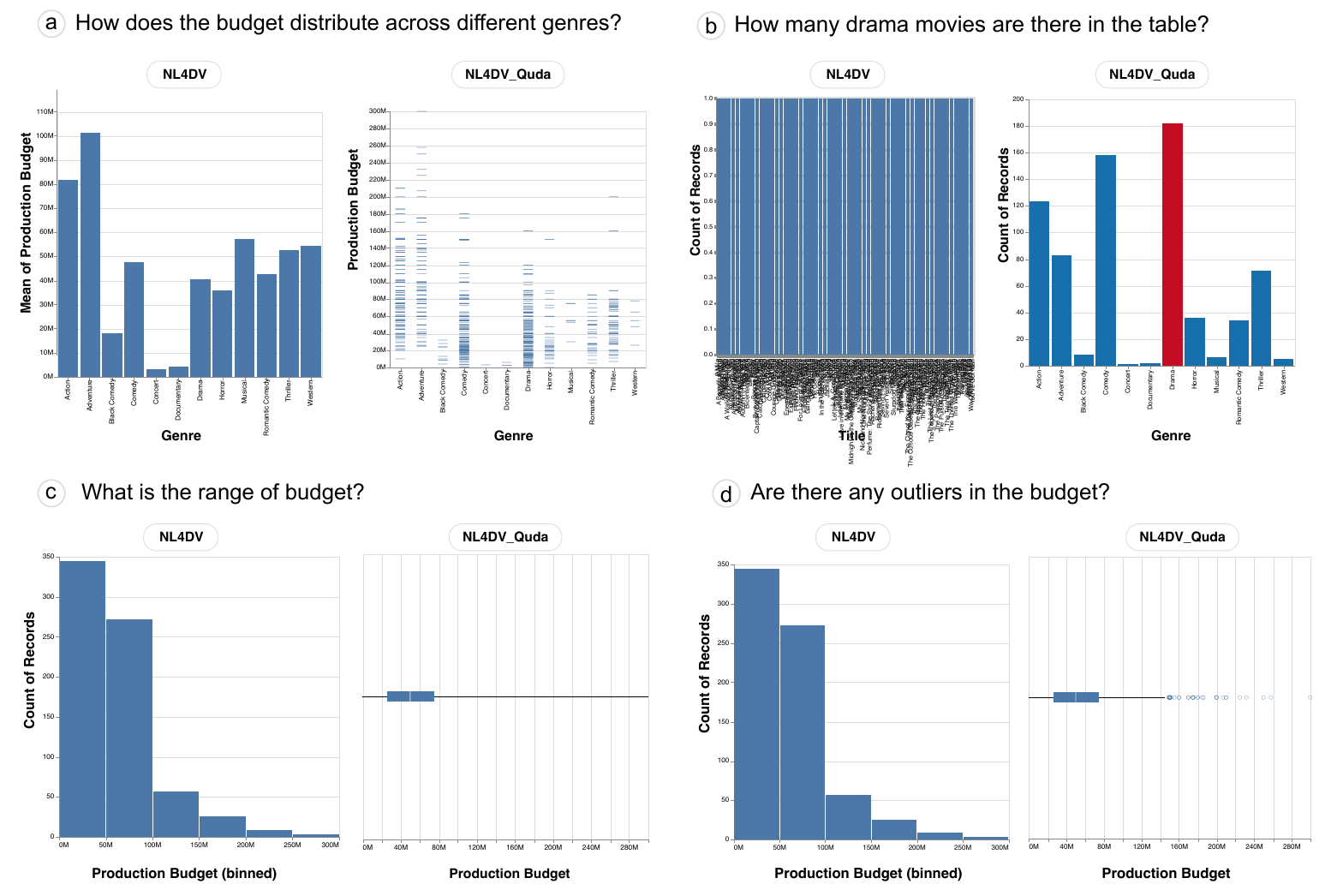}
    \caption{Four examples showcase that NL4DV\_\name{} can improve NL4DV in identifying tasks with higher precision  (a and b) and supporting more analytic tasks (c and d).}
    \label{fig:nl4dv}
\end{figure}

\subsubsection{Case Study}
We demonstrate that \name{} can improve the robustness of NL4DV in task inference.
We compare the results generated by NL4DV and the augmented library (named NL4DV\_\name{}) using a movie dataset.
Admittedly, NL4DV can generate reasonable visualizations in most cases. 
However, it fails when tasks are not correctly detected.
In the first example, we demonstrate that incorrect task inference leads to ineffective visualizations. 
Figure~\ref{fig:nl4dv}(a) shows the results of \textit{``How does the budget distribute across different genres?''}
NL4DV identifies the sentence as \textit{Derived Value} according to the type of attributes, and calculates the mean budget for each genre.
However, the mean budget cannot characterize budget distribution for each genre.
NL4DV fails to detect the query as \textit{Distribution} because ``distribute'' is not included in the task keywords.
NL4DV\_\name{}, on the other hand, detects the task as \textit{Distribution} and responds to the query using a strip plot.
We observe that the budget of action movies distributes evenly between 20 million and 150 million.
While most comedy movies have a budget of around 20 million.

The second example shows that NL4DV may miss tasks associated with a query. 
The query, \textit{``How many Drama movies are there in this table?''} belongs to \textit{Derived Value} and \textit{Filter} because drama movies should be filtered from all genres and the amount is derived through a count operator.
The attribute inference module of NL4DV detects one data value (drama) only.
Hence, NL4DV merely identifies it as \textit{Filter} and shows a bar chart with severe visual clutter (Figure~\ref{fig:nl4dv}(b)).
On the other hand, NL4DV\_\name{} correctly recognizes both tasks. 
It shows a bar chart with count operator to respond to the \textit{Derived Value} task, and further highlights drama movies to \textit{Filter}.



Thirdly, we illustrate that NL4DV\_\name{} supports a broader range of analytic tasks compared to NL4DV, including \textit{Determine Range}, \textit{Retrieve Value}, \textit{Sort}, and \textit{Find Anomalies}. 
Taking the query \textit{``what is the range of budget''} as an example, NL4DV detects it as \textit{Distribution} based on the type of data attribute, and shows a histogram for the production budget (Figure~\ref{fig:nl4dv}(c)).
However, the binning operation hinders users from identifying the minimum and maximum value.
Similarly, NL4DV detects the query, \textit{``Are there any outliers in the budget''} as distribution and depicts a histogram by default (Figure~\ref{fig:nl4dv}(d)).
The histogram, however, fails to reveal movies with an abnormal budget.
NL4DV\_\name{} can correctly detect the tasks and present an appropriate visualization in each case.
As shown in Figure~\ref{fig:nl4dv}(c) and (d), NL4DV\_\name{} shows boxplots with different parameter settings to demonstrate the range and outliers, respectively.

To conclude, due to the defects in task inference, NL4DV may result in ineffective visualizations.
The model trained on \name{}, on the other hand, is more robust in multi-label task classification.
Further, the model supports more analytic tasks which extends NL4DV to a broader range of usage scenarios. 

\subsection{Query Recommendation for V-NLIs}
Some V-NLIs, such as Eviza~\cite{Setlur2016}, Articulate~\cite{Sun2010}, and FlowSense~\cite{Yu2020}, have embedded template-based auto-completion to improve user experience~\cite{Gao2015}.
That is, after a user query is partially completed, the interface provides hints for possible queries.
Though useful, these techniques are tailored to specific applications and grammars, and can hardly be generalized to other V-NLIs.
Query recommendation techniques in search engine~\cite{BaezaYates2005,Zhang2006,Anagnostopoulos2010} suggest a list of related queries based on previously issued queries, which are independent of applications and have wide usage scenarios.
With the help of \name{}, we aim to develop a query recommendation technique for V-NLIs, which recommends semantically meaningful queries based on user input, data table, and \name{}.

\subsubsection{Technique Overview}
We first present the data annotation step that adapts \name{} to our query recommendation technique.
Then, we introduce our technique comprised of three modules: (1) \textit{candidate generation}, (2) \textit{candidate refinement}, and (3) \textit{interactive interface}.

In data annotation, we manually label $290$ expert-generated queries for demonstration, and plan to annotate all $14,035$ queries in \name{} in future research.
For each query, we distinguish phrases or words that are data attributes and data values. 
We further annotate the type of attributes and values as quantitative, nominal, ordinal, and temporal.
For example, given a query \textit{``what is the \dattribute{birth} distribution of the different \dattribute{districts} in \dvalue{2017}''}, we annotate ``birth'' as a data attribute with type quantitative, ``districts'' as a data attributes with type nominal, and ``2017'' as a data value with type temporal.

The three modules of our technique interact with each other.
Given a user input, the goal of the \textit{candidate generation} module is to identify candidate queries among $290$ queries that are semantically equivalent to the input.
We borrow universal sentence encoder~\cite{Cer2018} to vectorize $290$ queries and the user input, and compute the cosine similarity between them.
We then accept the top $5$ candidate queries  that are semantically ``close'' to the user input.
Since candidate queries do not relate to the data table, the recommendation, however, may not be meaningful for users.
The \textit{candidate refinement} module aims to refine the candidate queries by replacing data attributes (or values) in candidate queries with those in the data table.
We setup two criteria in the replacement: 1) the type of data attributes (or values) should be the same, and 2) the semantics should be as similar as possible.
Finally, we borrow NL-driven Vega-Lite editor~\cite{Narechania2020} as a testbed to showcase our query recommendation technique.
The interface is a Vega-Lite editor~\cite{Satyanarayan2017} enhanced with NL4DV which is a python-based NLI toolkit, which consists of an input bar, an editor for Vega-Lite specification, a panel for visualization results, and a panel for design alternatives.
When a user stops typing in the input bar, the input text is sent to the \textit{candidate generation} module.
After candidate queries are tuned by the \textit{candidate refinement} module, the results are shown in a dropdown menu.

\subsubsection{Usage Scenario}

\begin{figure}[tb]
    \centering
    \includegraphics[width=\linewidth]{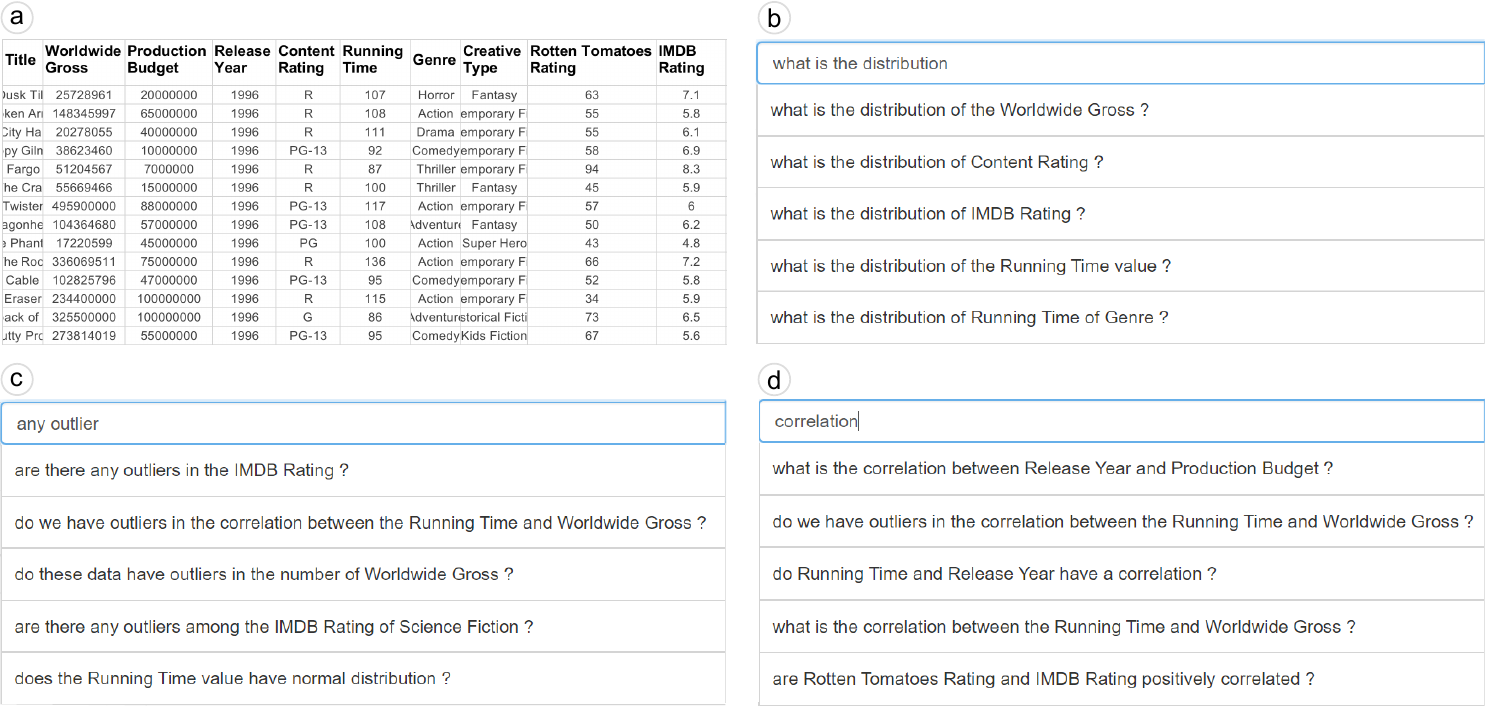}
    \caption{Given a data table (a), our technique recommends queries for three user inputs (b, c, d).}
    \label{fig:recommend}
\end{figure}

Suppose Jane is a fan of Hollywood movies, and she exhibits considerable interest in analyzing a movie dataset containing $709$ records with $10$ fields, including Running Time, Title, Genre, etc (Figure~\ref{fig:recommend}(a)). 
After selecting the Movies dataset, Jane wants to see how IMDB ratings are distributed. 
Then she types \textit{``what is the distribution''}.
A dropdown menu is then displayed showing the top $5$ queries recommended by our technique (Figure~\ref{fig:recommend}(b)).
The recommendation accelerates the input process, and Jane simply selects the third query for investigation.
Next, Jane wants to find abnormal movies in the dataset.
However, she has no idea where to start because there are  multiple ways to define whether a movie is abnormal.
She types \textit{``any outlier''} and pauses.
Our technique returns five queries to inspire her with the possible definition (Figure~\ref{fig:recommend}(c)).
The first sentence catches her eye.
IMDB rating is a commonly-used criterion to evaluate the quality of a movie.
She then chooses the first sentence for detailed inspection.
Finally, Jane wonders whether there are correlations between any data attributes, and types \textit{``correlation''}'.
The dropdown menu shows five possible results (Figure~\ref{fig:recommend}(d)).
Two of them suggest the correlation between release year and production budget.
Hence, Jane decides to start with the first query.
Finally, she concludes, \textit{``The query recommendation not only accelerates my exploration process, but also give me valuable suggestions when I do not exactly know what to see.''}

\section{Discussions}
\name{} is the first attempt to construct a large-scale corpus for V-NLIs. 
And we have demonstrated that \name{} can benefit the construction of V-NLIs in various scenarios.
Admittedly, some limitations exist in the current version of the corpus.
First, according to the six characteristics discussed in Section~\ref{sec:def}, our corpus covers a limited scope of queries. 
For example, \name{} does not include contextual queries that frequently appear in V-NLIs and systems with multimodal interaction~\cite{Srinivasan2018}. 
\name{} is a start for corpus collection in the domain of V-NLI.
We plan to construct more corpora to facilitate the research and development of V-NLIs.

Second, data tables have a strong influence on queries. 
Data tables with various domains would diversify queries in semantics and syntax.
Our corpus is constructed with $36$ data tables in $11$ domains. 
Though enumerating all data tables is infeasible, we should collect queries based on the most representative ones and cover as many domains as possible.
In future research, we will explore the datasets used in the domain of visual data analytics, and extend \name{} with more data tables involved. 

Third, paraphrase validation ensures the quality of crowd-authored queries. 
However, our approach requires extensive crowd force for pairwise comparisons, which would not be scalable if the number of paraphrases increases.
As \name{} is a large-scale corpus with human validation, we plan to train a machine learning model on our corpus to distinguish paraphrases with low quality.
Then the scalability issue can be alleviated.

\section{Conclusion and Future Work}
In this work, we present a large-scale corpus named \name{}, to help V-NLIs categorize analytic tasks by training and benchmarking machine/deep learning techniques.
\name{} contains $14,035$ queries labeled with some of $10$ low-level analytic tasks.
To construct \name{}, we first collect $920$ expert queries by recruiting $20$ data analysts.
Then, we employ crowd force to author $20$  paraphrases for each expert query.
Next, to ensure the quality of paraphrases, we borrow the crowd to collect the pairwise comparison dataset and compute semantic equivalence scores using relative attribute~\cite{Parikh2011}. 
Finally, paraphrases with low scores are filtered out.
We present three experiments to illustrate the usefulness and significance of \name{} in different usage scenarios. 

In the future, we plan to explore more research opportunities based on \name{}.
First, we plan to continuously expand the scope of \name{} in several aspects, such as  
1) supporting abstract queries, 2) including queries that have high composition level, 3) including queries with actions, and 4) involve more data tables and experts.
Second, we plan to improve the process of corpus construction to balance data quality and cost.
The cost for expert query collection and paraphrase generation is reasonable. 
However, the process of paraphrase validation is expensive, and the cost will increase quadratically with the number of paraphrases.
With the help of \name{}, we plan to train a classification model to investigate scalable approaches for quality control.
Third, a promising direction of V-NLI is the support of multi-turn conversations, which involves the inference of contextual information of a query.
We plan to investigate how \name{} can support multi-turn conversations.

\bibliographystyle{ACM-Reference-Format}
\bibliography{template}

\end{document}